\documentclass[
]{ceurart}

\usepackage{listings}
\usepackage{todonotes}

\usepackage{xcolor}

\usepackage{graphicx}
\usepackage{cleveref}
\usepackage{enumitem}
\usepackage{multirow}

\lstdefinestyle{codeblock}{
    basicstyle=\ttfamily\small,
    keywordstyle=\color{blue!70!black},
    commentstyle=\color{gray!80},
    stringstyle=\color{green!50!black},
    showstringspaces=false,
    breaklines=true,
    frame=single,
    framerule=0.2pt,
    backgroundcolor=\color{gray!5}
}

\begin{document}

\copyrightyear{2025}
\copyrightclause{Copyright for this paper by its authors.
  Use permitted under Creative Commons License Attribution 4.0
  International (CC BY 4.0).}

\conference{LLMS4KGOE @ ESWC'26}

\title{Towards Automated Ontology Generation from Unstructured Text: A Multi-Agent LLM Approach}

\author{Abid Talukder}[%
email=taluka@rpi.edu,
]

\author{Maruf Ahmed Mridul}[%
email=mridum@rpi.edu,
]

\author{Oshani Seneviratne}[%
email=senevo@rpi.edu,
]

\address{Rensselaer Polytechnic Institute, Troy, New York, United States}

\begin{abstract}
Automatically generating formal ontologies from unstructured natural language remains a central challenge in knowledge engineering. While large language models (LLMs) show promise, it remains unclear which architectural design choices drive generation quality and why current approaches fail. We present a controlled experimental study using domain-specific insurance contracts to investigate these questions.
We first establish a single-agent LLM baseline, identifying key failure modes such as poor Ontology Design Pattern compliance, structural redundancy, and ineffective iterative repair. We then introduce a multi-agent architecture that decomposes ontology construction into four artifact-driven roles: Domain Expert, Manager, Coder, and Quality Assurer.
We evaluate performance across architectural quality (via a panel of heterogeneous LLM judges) and functional usability (via competency question driven SPARQL evaluation with complementary retrieval augmented generation based assessment). Results show that the multi-agent approach significantly improves structural quality and modestly enhances queryability, with gains driven primarily by front-loaded planning.
These findings highlight planning-first, artifact-driven generation as a promising and more auditable path toward scalable automated ontology engineering.
\end{abstract}


\maketitle

\section{Introduction}
Ontologies, built upon formal languages like the Web Ontology Language (OWL), provide a robust framework for representing domain knowledge in a structured, machine-readable format, enabling applications such as automated reasoning, semantic search, and data integration. However, manual ontology development is labor-intensive, error-prone, and requires significant domain expertise, which is a bottleneck that limits the scalability of knowledge-based systems. The emergence of large language models (LLMs) has opened new avenues for automating this process, but key questions remain open: how reliably can LLMs generate formally valid, semantically rich ontologies from unstructured text, and what architectural choices actually drive quality?

This challenge is especially pronounced in domains where knowledge is embedded in complex natural language documents such as legal contracts, regulatory filings, technical specifications, where entities, relationships, and constraints must be inferred rather than directly read off a structured schema. Converting such unstructured text into a formal, queryable ontology demands not only language understanding but also adherence to modeling best practices that single-pass LLM generation frequently violates.

We investigate these questions through a controlled experimental study. Starting from a direct LLM-based generation approach as a baseline, we identify its systematic limitations and use them to motivate a multi-agent architecture in which ontology construction is decomposed into specialized, artifact-driven roles. By holding requirement extraction and evaluation constant across both approaches, we isolate the impact of the construction strategy itself. The study surfaces concrete failure modes like redundancy, context degradation, and unreliable iterative remediation that points toward the architectural improvements needed for a more robust next-generation pipeline.

\subsection{Key Contributions}
\begin{itemize}
\item An experimental study of LLM-driven ontology generation from unstructured domain text, comparing a direct generation baseline against a multi-agent architecture under controlled, identical conditions.
\item An automated CQ-driven evaluation framework that uses iterative, feedback-driven SPARQL query generation to measure the pragmatic usability of generated ontologies.
\item A diagnostic analysis of failure modes in LLM-based ontology generation, with concrete architectural insights that inform the design of future automated ontology engineering systems.
\end{itemize}

\subsection{Motivating Use Case: Ontology Construction from Life Insurance Policies}

Life insurance policies provide a representative example of the challenges involved in transforming complex natural language documents into formal ontologies. Insurance contracts contain dense legal language describing entities (e.g., \emph{policyholder}, \emph{beneficiary}), events (e.g., policy activation, reinstatement, claim submission), financial constraints (e.g., coverage limits and premium amounts), and temporal conditions (e.g., exclusion periods). These elements are often expressed implicitly across multiple clauses and require careful interpretation to model correctly.
As such, insurance contracts exhibit rich relational, temporal, and constraint-heavy structures, making them a representative stress test for ontology generation. While domain-specific, the challenges (multi-entity reasoning, temporal constraints, financial modeling) generalize to other legal and technical documents.

Consider a typical clause from a life insurance contract:

\begin{quote}
``The beneficiary is entitled to a one-time payment of \$50{,}000 upon the policyholder's death.''
\end{quote}

While the statement appears straightforward in natural language, representing it formally in an ontology requires modeling multiple interacting concepts: the \texttt{PaymentEvent}, the \texttt{Beneficiary} role, the triggering \texttt{DeathEvent} of the policyholder, and the associated monetary value and currency constraints. Additionally, this representation must support downstream reasoning tasks, such as answering competency questions (CQs) like:

\begin{quote}
\emph{What payment is the beneficiary entitled to upon the death of the policyholder?}
\end{quote}

Traditional ontology engineering would require a domain expert and a knowledge engineer to manually identify these concepts, apply appropriate ontology design patterns (ODPs), and encode the resulting structure in OWL. When applied across an entire contract, often spanning dozens of pages and hundreds of clauses, this process becomes labor-intensive and difficult to scale.

In this work, we use life insurance contracts as a running example to investigate how LLMs can assist in automating this process. Specifically, we examine how an LLM-driven pipeline can transform policy text into a formal ontology by first deriving CQs from contract clauses and then generating structured representations of the relevant entities, events, and constraints. Throughout the paper, examples and experiments are drawn from publicly available insurance policy documents to illustrate both the capabilities and limitations of automated ontology generation.

\section{Related Work}

The use of LLMs to automate ontology engineering is a growing research area, as the traditional development process is complex and requires significant human expertise~\cite{lippolis2025ontology, saeedizade2024navigating, bakker2024ontology, norouzi2025ontology, babaei2023llms4ol}. However, the existing work primarily focuses on (i) single-pass ontology generation, (ii) subtask-level assistance (e.g., taxonomy construction), or (iii) schema-guided population. In contrast, our work explicitly investigates construction strategy as the independent variable, introducing an artifact-driven multi-agent decomposition and isolating its impact under controlled evaluation.

Lippolis et al. investigated the potential of LLMs to create OWL ontology drafts from user stories and CQs, demonstrating that new prompting techniques could outperform novice human modelers~\cite{lippolis2025ontology}. Similarly, Saeedizade et al. explored the use of LLMs for generating OWL modeling suggestions from ontological requirements, noting that advanced models could provide suggestions of sufficient quality to assist human modelers~\cite{saeedizade2024navigating}. Kommineni et al. also explored a semi-automated pipeline for constructing knowledge graphs and ontologies using open-source LLMs, leveraging a judge LLM for evaluation~\cite{kommineni2024human}. While these studies show the promise of LLMs for initial ontology generation, a key challenge remains in ensuring the formal correctness and reliability of the output, a necessity for high-stakes domains.

To address the limitations of one-shot generation, other works have focused on specific subtasks of ontology learning. Babaei et al. introduced \textit{LLMs4OL} to evaluate LLMs on core tasks like term typing, concluding that fine-tuned models could assist in ontology construction~\cite{babaei2023llms4ol}. Similarly, Lo et al. proposed \textit{OLLM} for building the taxonomic backbone of an ontology~\cite{lo2024end}. While these advances are promising, issues such as hallucination and inconsistent output persist, as observed by Bakker et al.~\cite{bakker2024ontology}. Our work extends these efforts by introducing a novel, cooperative multi-agent architecture that enforces a structured, artifact-driven workflow to mitigate these issues and ensure higher quality output. Norouzi et al. also investigated LLM effectiveness for ontology population, but from a pre-defined schema, achieving high triple extraction rates under guidance~\cite{norouzi2025ontology}.

The challenges of generating reliable ontologies become even more pronounced when applied to domain-specific knowledge. Chowdhury et al. and Abolhasani et al. developed frameworks for generating ontologies from text in cultural and technical domains, respectively, using methods like adaptive Chain-of-Thought (CoT) to ensure alignment with user needs~\cite{chowdhury2025automated, abolhasani2024leveraging}. A separate but related line of work has focused on applying ontology engineering to the legal and financial sectors. Previous research in the insurance domain has typically relied on human experts to define the ontology and its rules, with work by Charalambous et al., Ahaggach et al., Naqvi et al., and Okikiola et al. focusing on either populating or utilizing a pre-existing schema for applications like pricing and healthcare management~\cite{charalambous2022analyzing, ahaggach2023information, naqvi12023ontology, okikiola2020systematic}.

A critical component of building executable knowledge systems is the formalization of rules and the use of CQs to guide development. CQs are a well-established method for guiding and evaluating ontologies~\cite{lippolis2025ontology, bezerra2013evaluating, araujo2016data, article, feng2024ontologygroundedautomaticknowledgegraph}, but their use in a fully automated generation and evaluation process is an active research area. Similarly, the Semantic Web Rule Language (SWRL)~\cite{horrocks2004swrl} has been applied to various domains for rule formalization, such as legal reasoning in criminal law by El Ghosh et al. and for managing informed consent permissions by Amith et al.~\cite{el2017towards, amith2022expressing}. However, these works rely on manually authored rules and schemas rather than deriving structure from unstructured text. Our work addresses this gap by investigating how LLM-driven ontology generation can be made to produce higher-quality, formally valid ontologies from unstructured domain text: identifying the limitations of direct generation, exploring a multi-agent architecture as a means to address them, and surfacing concrete directions for further improvement.

\section{Methodology}

Our approach treats ontology generation from unstructured text as a pipeline problem decomposed into three stages: requirement extraction, ontology construction, and automated validation. Rather than proposing a finished system, we report on a structured experimental investigation. We first establish a direct LLM-based generation baseline, identify its limitations, and then design and evaluate a multi-agent extension that attempts to address those limitations. The goal is to understand what architectural  choices actually drive quality improvements, and to surface the failure modes that must be overcome in future work.

Formally, both of our experiments implement a function $\mathcal{F}$ that transforms an unstructured document $D$ into a domain ontology $\mathcal{O}$ together with a set of evaluation artifacts \( \mathcal{E} \): \( \mathcal{F}(D) \to (\mathcal{O}, \mathcal{E}) \).

The transformation proceeds through three stages: (1) CQ generation, which derives structured requirements from the source text; (2) ontology generation, which constructs an OWL/TTL ontology guided by those requirements; and (3) automated SPARQL-based evaluation, which assesses whether the ontology can answer each requirement.

\subsection{Competency Question (CQ) Generation}

In ontology engineering, CQs serve as functional requirements, precisely defining the scope of knowledge the ontology must represent and the queries it must answer ~\cite{lippolis2025ontology, article, feng2024ontologygroundedautomaticknowledgegraph}. In our methodology, they fulfill three key objectives: first, they specify the minimal set of facts, relationships, and constraints required for the ontology; second, they enforce the use of ODPs such as Event Reification, Participation, Quantities and Qualities, Time Instants and Intervals, or Situations and States~\cite{shimizu2019modl,guizzardi2022ufo}, ensuring adherence to modeling best practices; and third, they are formulated to be directly expressible as SPARQL queries without re-interpretation, which enables objective, machine-verifiable evaluation. By binding each CQ to specific ODPs, our approach compels the ontology generation model to produce semantically structured and logically consistent representations.

The methodology begins by processing contracts in a document-by-document, page-by-page manner. This approach grounds each CQ in a manageable textual context, which prevents context overflow in the language model and provides a fine-grained link between contract sections and the resulting CQs. For each page, the relevant text is extracted and passed to a controlled prompting environment. Here, an LLM is instructed via a structured prompt to output results in a fixed schema containing exactly two fields: a precisely formulated CQ and its corresponding \textit{expected answer}. This schema design enforces clarity, reduces ambiguity, and ensures that the generated CQs remain directly convertible into SPARQL queries without additional interpretation. While the use of ODPs informs the overall CQ formulation process, no explicit ODP annotations or auxiliary metadata are embedded in the CQ outputs.

The process of CQ generation can be formalized as a function $\mathcal{G}_{CQ}: \mathcal{T} \to \mathcal{C}$, where a given text page $\mathcal{T} \in \mathbb{T}$ is mapped to a set of CQs $\mathcal{C} = \{CQ_1, CQ_2, \dots, CQ_m\}$. The LLM's output for each CQ strictly adheres to the two-field schema, ensuring structured, unambiguous, and machine-interpretable data generation.

For example, given the contract clause, \textit{`Reinstatement: The process of restoring a lapsed Policy to active status as outlined in this Policy.'}, the LLM generates the example CQ listed in \Cref{lst:cq}. This decomposition enforces a rich, reified model that is more extensible and capable of answering nuanced queries than a flat set of predicates. 

\begin{lstlisting}[style=codeblock,label=lst:cq, caption={Example CQ}]
{
  "competency_question": "What is the name of the event when a lapsed policy is restored to active status?",
  "expected_answer": "Reinstatement"
}
\end{lstlisting}

A second example illustrates the use of time-related ODPs. Given the text \textit{`The policy contains a suicide exclusion period of two years from the effective date'}, a CQ would be generated to model the temporal restriction: \textit{`What is the duration of the `Suicide Exclusion Period' event?'}. The explicit mention of \textit{duration} in the CQ directly invokes the temporal ODP, signaling the ontology generation process to represent both the event and its associated time interval. Without such alignment to the ODP, the CQ might instead prompt a model to capture only an absolute end date, omitting the interval structure and thereby limiting the ontology's ability to support temporal reasoning, comparative date queries, or duration-based constraints.

The structured, machine-readable nature of the CQs serves three purposes: (i) it guides the ontology generator, (ii) establishes a fixed benchmark for evaluation, and (iii) preserves the provenance of each CQ for human auditing. Ultimately, this CQ generation approach defines not only what the ontology must represent but also how it should be structured to represent it, reducing modeling ambiguity and laying the groundwork for automated evaluation.

\subsection{Baseline Direct Ontology Generation Method}

\begin{figure}[t]
    \centering
    \includegraphics[width=0.8\linewidth]{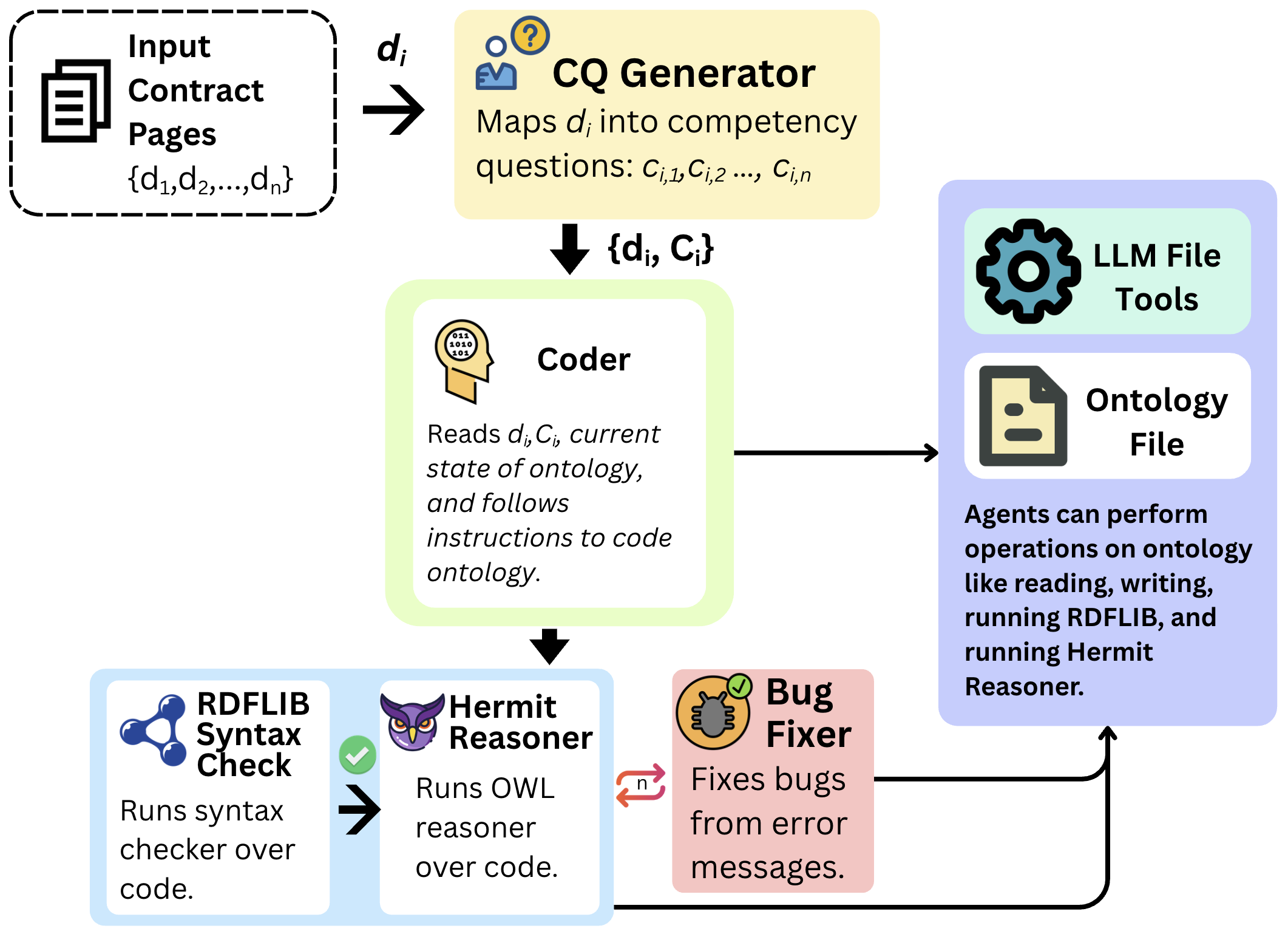}
    \caption{Workflow of the baseline ontology generation method}
    \label{fig:baseline_method}
\end{figure}

Our first experiment establishes a controlled, minimal-interaction scenario against which we can measure the benefits of a more integrated, iterative, and feedback-driven approach. This baseline allows us to quantitatively demonstrate what is gained by introducing the more complex multi-agent framework. This method is a simple, functional version of the ontology generation pipeline, where each major stage operates in isolation. This approach is characterized by a lack of feedback or iterative refinement between stages; each stage runs once and its output becomes the fixed input for the next. As illustrated in Figure \ref{fig:baseline_method}, the workflow is comprised of two discrete, sequential stages:
$\text{Generate}(\mathcal{T}, \mathcal{C}) \to \mathcal{O}_{raw} \to \text{Validate}(\mathcal{O}_{raw}) \to \mathcal{O}_{final}$,
where the output of each stage is the fixed input for the next, with no opportunity for iterative refinement based on validation results.

\paragraph{Ontology Generation:}
This stage constructs the ontology in Turtle (TTL) format directly from the contract text and the CQs. The LLM agent is instructed to represent all necessary concepts and relationships, applying ODPs like event reification and participation modeling, and extending the TTL file incrementally. This agent operates without any knowledge of downstream validation results. The generator processes the contract text page-by-page, receiving the text for that page and the corresponding CQs. It is tasked with representing all concepts, relationships, and constraints needed to answer the CQs and with applying ODPs to the model. The agent is instructed to extend the ontology by appending or editing triples in a persistent TTL file. Crucially, even if the generated ontology contains modeling errors, redundant entities, or incomplete structures, these will not be addressed in the generation stage once it has concluded. For example, if the agent fails to define a required class for a concept like `premium payment', this omission will carry forward into the next stage.
\begin{wrapfigure}{r}{0.5\textwidth}
  \vspace{-22pt} 
  \begin{minipage}{\linewidth}
\begin{lstlisting}[style=codeblock,label=lst:modeling-error, caption={Example Modeling Error}]
ex:PolicyPayment rdf:type owl:Class .
ex:hasAmount rdfs:domain ex:PolicyPayment ;
    rdfs:range xsd:decimal .
ex:policy1 ex:hasPayment ex:payment1 .
ex:payment1 ex:hasAmount 5000 .
\end{lstlisting}
\end{minipage}
  \vspace{-20pt} 
\end{wrapfigure}
To mitigate common LLM modeling errors, we incorporated a curated set of prompting examples adapted from the \textit{Memoryless CQbyCQ} and \textit{Ontogenia} techniques \cite{lippolis2025ontology,Lippolis2024OntogeniaOG}, which enumerates frequent ontology generation and design mistakes, and ODPs, using them to guide the LLM toward more accurate and semantically consistent outputs. 

Refer to \Cref{lst:modeling-error}, for concrete example of such a modeling mistake, which omits the unit for the monetary amount, making the CQ `What is the monetary amount the beneficiary is entitled to?' unanswerable in a semantically precise way.
These instantiated individuals are introduced solely for evaluation purposes and should be interpreted as synthetic test cases rather than part of the core ontology design. Their role is to enable SPARQL-based validation by exercising the underlying schema, ensuring that the defined classes and properties can support executable query patterns aligned with the CQs.

\paragraph{Ontology Validation and Bug Fixing:} This is a standalone process that takes the ontology file from the generator. This process has no interaction with the generator and cannot request a re-run of the generation phase. The validation phase consists of two checks: a \textbf{Syntax Check} using an RDF parser \footnote{\url{https://rdflib.readthedocs.io/en/stable}} to ensure compliance with TTL syntax, and a \textbf{Semantic Consistency Check} using an OWL reasoner ~\cite{LAMY201711, Shearer2008HermiTA} to ensure logical coherence. If an error is detected at either stage, the file is routed to a bug-fixing agent that reads only the relevant sections associated with the reported error. For instance, if the syntax checker finds a malformed triple, the bug-fixing agent will read the specific lines containing the error, apply a minimal, targeted fix (e.g., correcting a missing period or prefix), and then return the modified file to the start of the validation process. This cycle repeats until the ontology passes both checks or a maximum number of attempts (set to 1000 by default) is reached. This decoupled workflow means the \emph{Bug Fixer} agent never reinterprets the contract or CQs, but only resolves formal errors. It does not address structural or conceptual omissions made by the generator, such as missing entities or relationships. Refer to \Cref{lst:fixed,lst:original} for an example of a syntax error and fix, where the fix addresses the syntax, but it doesn't change the underlying conceptual model.

\begin{center}
\begin{minipage}{0.48\textwidth}
\begin{lstlisting}[style=codeblock, caption={Original, malformed TTL fragment}, label=lst:original]
ex:policyOwner rdf:type owl:Class
ex:premiumPayment rdf:type owl:Class .
\end{lstlisting}
\end{minipage}
\hfill
\begin{minipage}{0.48\textwidth}
\begin{lstlisting}[style=codeblock, caption={Bug-fixing agent correction},label=lst:fixed]
ex:policyOwner rdf:type owl:Class ;
ex:premiumPayment rdf:type owl:Class .
\end{lstlisting}
\end{minipage}
\end{center}

\subsection{Multi-Agent Generator (Domain Expert $\to$ Manager $\to$ Coder $\to$ Quality Assurance)}
The multi-agent generator is designed to enhance modeling fidelity and reduce rework by decomposing the ontology construction process into four specialized, cooperative roles: (i) \emph{Domain Expert}, (ii) \emph{Manager}, (iii) \emph{Coder}, and (iv) \emph{Quality Assurer}. These agents operate on a single contract page at a time, but they coordinate through shared, structured outputs that preserve requirements, design intent, and quality expectations. This design enforces planning before coding, governs modeling with ODPs, and incorporates architectural quality assurance that complements formal checks.

\begin{figure}[t]
    \centering
    \includegraphics[width=0.75\linewidth]{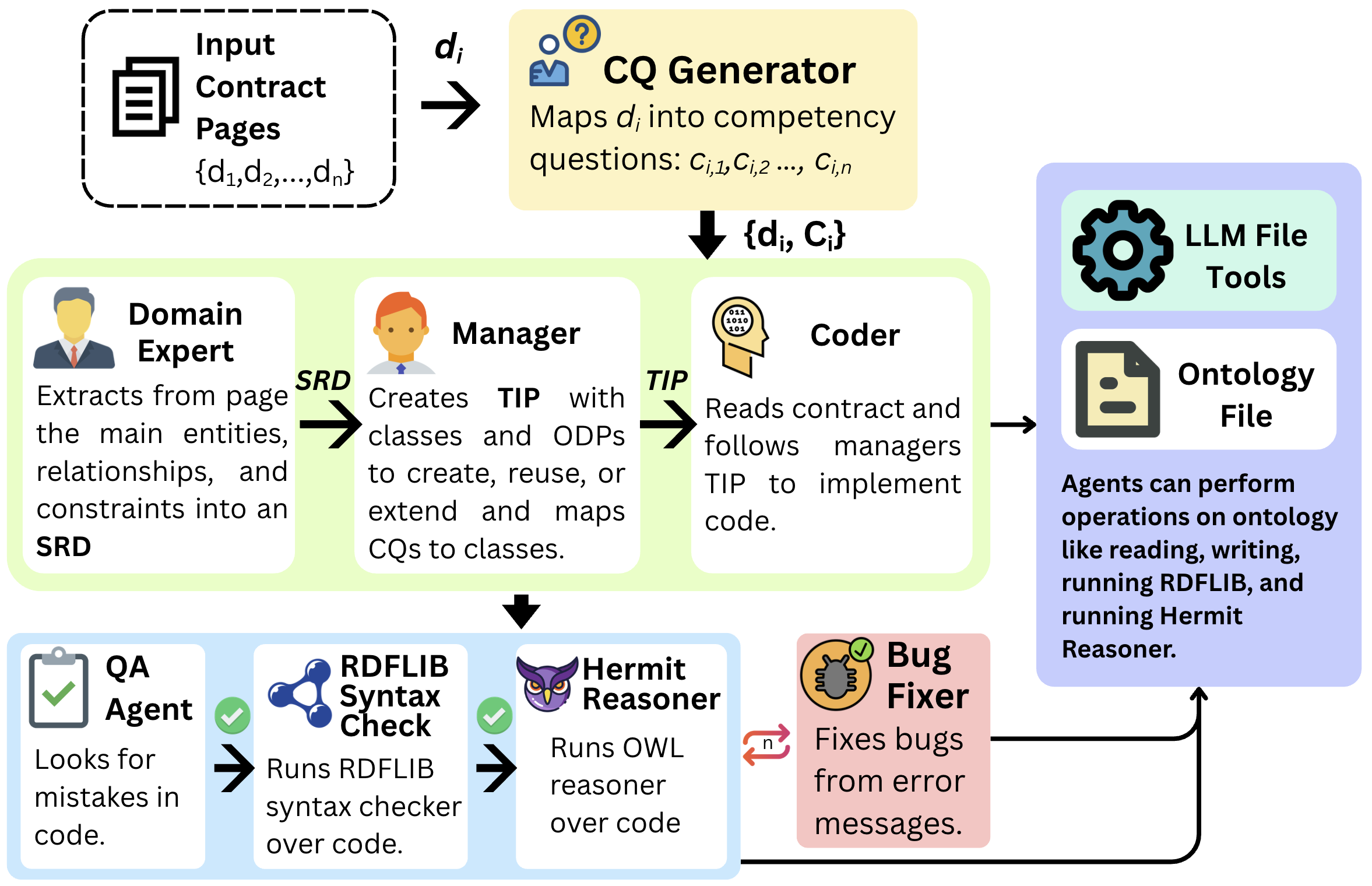}
    \caption{Workflow of the multi-agent ontology generation method (our primary contribution)}
    \label{fig:multi_agent_method}
\end{figure}

The multi-agent pipeline (see Figure \ref{fig:multi_agent_method}) can be modeled as a directed acyclic graph (DAG) of agents, where each agent's output is a structured artifact that serves as the input for the next. This formalizes the artifact-driven hand-off principle, with the generation of the Semantic Requirements Document (SRD), and the Technical Implementation Plan (TIP) as its intermediate steps.

$$\mathcal{G}_{multi-agent}: (T, \mathcal{C}) \to SRD \to TIP \to \Delta\mathcal{O} \to \mathcal{O}_{final}$$
The bounded iteration in the QA stage can be modeled as a loop: $\text{QA}(\Delta\mathcal{O}) \to \text{fix}(\Delta\mathcal{O}) \to \dots \text{ (up to } n \text{ iterations)}$, where $n$ is set to 1000 by default.

\subsubsection{Design Principles}

\begin{itemize}[leftmargin=*]
    \item \textbf{Separation of concerns} allows each agent to focus on a single competency, semantics, architectural planning, code realization, or quality assurance, which reduces cognitive load.
    \item \textbf{Artifact-driven hand-offs} ensure that every stage produces a consumable, specified artifact (e.g., a requirements document, an implementation plan, code edits, or QA findings), minimizing ambiguity and preserving traceability.
    \item \textbf{ODP governance} is enforced in the planning and QA stages, ensuring the coder implements a predefined structure rather than improvising one.
    \item \textbf{CQ alignment} is maintained throughout the entire process, as CQs are carried into every stage as requirements, alignment matrices, acceptance criteria, and review checkpoints.
    \item \textbf{Bounded iteration} allows for local cycles of remediation to address concrete issues, but these are capped to prevent endless loops.
\end{itemize}

\subsubsection{Agent Roles}

\paragraph{Role 1: Domain Expert:}

The Domain Expert agent's objective is to extract a precise, contract-grounded SRD that enumerates the concepts, relationships, temporal constraints, and business rules necessary to answer the CQs. 

To achieve this, the agent first reads the page text and its CQs holistically. It then normalizes domain signals into compact lists, identifying key concepts labeled, such as \texttt{Entity}, \texttt{Role}, \texttt{Event}, \texttt{State}, \texttt{Attribute}, \texttt{TimeInterval}, and \texttt{MonetaryAmount}. It also captures relationships with optional qualifiers (e.g., triggers or exceptions), business rules, temporal and monetary constraints, and state machine fragments for policy lifecycles. 

All generated elements are tied to their corresponding CQs to ensure alignment. The necessity of this step stems from the observation that 
LLMs frequently conflate domain-specific linguistic constructs with ontological primitives, leading to misaligned class and property definitions.
By forcing a requirements-first articulation of the domain into a compact, auditable document, the SRD reduces the chance of architectural drift in later stages. The agent is explicitly instructed to avoid inventing new information and to prefer canonical terminology implicitly supported by the text.
The output of the Domain Expert is a structured SRD, and it serves as a formal contract for the next agent in the pipeline. For example, the SRD in \Cref{lst:domain-expert-output} is the output for the clause \textit{`The Beneficiary is entitled to a one-time payment of \$50,000 upon the policyholder's death'}.

\begin{lstlisting}[style=codeblock,label=lst:domain-expert-output, caption={Example Domain Expert Output}]   
[{
    "cq_id": "CQ-1234",
    "entities": [
      {"name": "Beneficiary", "type": "Role"},
      {"name": "Policyholder", "type": "Entity"},
      {"name": "Payment", "type": "Event"}
    ],
    "relationships": [
      {"source": "Beneficiary", "target": "Payment", "label": "entitledTo"},
      {"source": "Payment", "target": "Policyholder", "label": "triggeredByDeathOf"}
    ],
    "constraints": [
      {"entity": "Payment", "type": "MonetaryAmount", "value": "50000", "unit": "USD"}
    ]
}]
\end{lstlisting}

\begin{lstlisting}[style=codeblock,label=lst:manager-tip, caption={Manager's Technical Implementation Plan (TIP) Prompt}]
### 1) Conceptual Model & Pattern Guidance
**Concept:** `ex:PaymentEvent`
**Represents:** A one-time transaction event triggered by the policyholder's death, resulting in a payout to the beneficiary.
**Modeling Approach (ODPs):** Use the **Event** pattern to represent the payment itself. Use the **Quantity and Units** pattern to encapsulate the $50,000 constraint so the currency and amount are kept together. Use **AgentRole** to decouple the person from their function as a Beneficiary.
**Key Relationships:** Acts as a subclass of `ex:Event`. Connects to the entitled `foaf:Person` via `ex:hasEntitledParty`. Connects to the financial constraint via `ex:hasMonetaryAmount`.

**Concept:** `ex:MonetaryAmount`
**Represents:** The structured financial value of the payment.
**Modeling Approach (ODPs):** Use the **Description and Details (Quantities and Units)** pattern to avoid primitive string typing for money. 
**Key Relationships:** Connects to `xsd:decimal` via `ex:hasValue` and `xsd:string` via `ex:hasCurrency`.

### 2) Reuse & Extension Plan
- **Extend/Reuse:** Reuse `ex:Event` as the parent class for the payment. Reuse `foaf:Person` to represent the underlying entities fulfilling the Policyholder and Beneficiary roles.
- **Create New:** Declare `ex:PaymentEvent` and `ex:MonetaryAmount` as new classes. Declare `ex:hasEntitledParty` and `ex:hasMonetaryAmount` as object properties. Declare `ex:hasValue` and `ex:hasCurrency` as data properties.

### 3) Competency Question Alignment
**Question:** What payment is the Beneficiary entitled to upon the death of the Policyholder?
**Answer:** The Beneficiary is entitled to a one-time payment event totaling 50,000 USD.
**Classes:** `ex:PaymentEvent', `ex:MonetaryAmount', `foaf:Person`
**ODP:** Event, Quantities and Units
\end{lstlisting}

\paragraph{Role 2: Manager:}

The Manager agent's objective is to translate the SRD into a build-ready TIP that codifies modeling choices before any code is written. The TIP is a structured, three-part document with the following sections:

\begin{enumerate}
    \item \textit{Conceptual Model \& Pattern Guidance} outlines the intent of the two to four most central concepts on the page and specifies the exact ODPs to use. To ensure pattern selection is grounded in established best practices rather than ad-hoc invention or having the LLM guess, the Manager is provided with a curated list of viable ODPs and their associated use cases, adapted from the MODL framework~\cite{shimizu2019modl}. 
    \item \textit{Reuse \& Extension Plan} enumerates existing classes and properties to be reused or subclassed, and lists the minimal set of new ones required with a clear rationale. 
    \item\textit{CQ Alignment} details for each CQ the classes involved, the primary ODP, and how it connects the classes to make the CQ answerable by SPARQL. This stage is critical because many ontology defects originate from coding without a clear architecture. The TIP front-loads these architectural decisions, making ODP use explicit and establishing a clear contract between the design intent and the final implementation, which also helps prevent the proliferation of redundant classes.
\end{enumerate}

As an example, for the SRD snippet in \Cref{lst:domain-expert-output}, the Manager produces a TIP with the guidance shown in \Cref{lst:manager-tip}.

\paragraph{Role 3: Coder:}
\begin{wrapfigure}{r}{0.4\textwidth}
  \vspace{-10pt} 
  \begin{minipage}{\linewidth}
\begin{lstlisting}[style=codeblock,label=lst:coder-output, caption={Example Output by the Coder}]  
ex:PaymentEvent a owl:Class ;
  rdfs:subClassOf ex:Event .
ex:hasEntitledParty a owl:ObjectProperty ;
  rdfs:domain ex:PaymentEvent ;
  rdfs:range foaf:Person .
ex:payment1 a ex:PaymentEvent ;
  ex:hasEntitledParty ex:beneficiary1 ;
  ex:hasMonetaryAmount [
    a ex:MonetaryAmount ;
    ex:hasValue "50000"^^xsd:decimal ;
    ex:hasCurrency "USD"  ] .
\end{lstlisting}
\end{minipage}
  \vspace{-15pt} 
\end{wrapfigure}

The Coder agent's objective is to implement the TIP by producing concrete TTL edits and instantiating facts from the contract page in a way that ensures the CQs are answerable. The agent's method involves reading the current ontology snapshot, which is provided with line numbers to encourage the reuse of existing constructs before adding new ones. It then implements classes, properties, restrictions, and reified event/role structures as directed by the TIP. For instance, the coder will separate numeric values from currency units, represent durations as intervals, and connect event participants through object properties.
As with the baseline method, prompting examples from the \textit{Memoryless CQbyCQ} and \textit{Ontogenia} techniques \cite{lippolis2025ontology,Lippolis2024OntogeniaOG} were incorporated to further reduce common modeling errors. The necessity of these steps lies in its ability to curb speculative elaboration and ensure continuity, as the implementation is tightly anchored to the pre-approved plan, producing changes that are both traceable and reviewable.
Based on the TIP, the Coder would generate TTL code edits that append or modify the ontology. \Cref{lst:coder-output} demonstrates a snippet of this generated code.
This code adheres to the specified ODPs and directly represents the facts needed to answer the CQ.

\paragraph{Role 4: Quality Assurer (QA):}
The QA agent's objective is to enforce quality through a three-gate sequence: (i) an architectural review, (ii) a syntax check, and (iii) a semantic consistency check, with a dedicated \textbf{QA-Coder} applying fixes as needed. 

The first gate, the \textbf{QA Architectural Review}, is performed by an LLM that judges the code's fidelity to the TIP and the original contract text. It identifies architectural violations, such as incorrect domains or ranges, missing classes, or misuse of ODPs. This review provides structured, line-anchored feedback (e.g., ``L123--L140: Participation inverted; event lacks agent link'') to guide remediation. 

The subsequent gates, \textbf{Syntax Verification} (using a TTL parser) and \textbf{Semantic Consistency Verification} (using an OWL reasoner ~\cite{LAMY201711, Shearer2008HermiTA}), catch formal errors. Unlike the baseline method, this QA stage is unique to our multi-agent framework because it is able to assess architectural quality by referencing the artifacts produced by prior agents, such as the \textbf{Manager}'s TIP and the \textbf{Coder}'s implementation. These artifacts are essential for verifying that the agents worked cohesively and adhered to a shared design intent, enabling the QA process to evaluate not only individual correctness but also cross-stage alignment and overall modeling fidelity. 

The \textbf{QA-Coder} then reads only the relevant slices of the ontology and applies precise, targeted edits to fix the issues, looping back through the gates within a bounded number of iterations to prevent infinite loops. 

\Cref{lst:qa-feedback} shows an example of a feedback message from the QA Review and the subsequent action by the QA-Coder.
The QA-Coder then receives this feedback and applies a targeted fix, ensuring the correct range is specified, and the process repeats.

\begin{lstlisting}[style=codeblock,label=lst:qa-feedback, caption={QA Review Feedback}]
{
  "file": "ontology.ttl",
  "description": "
  In L35-37, there is an architectural violation. Property 'ex:hasValue' lacks a typed range. Please specify 'xsd:decimal' as required by ODP Quantities."
}
\end{lstlisting}

\subsection{Evaluation Framework}
\label{sec:evaluation}

\begin{figure}[t]
    \centering
    \includegraphics[width=0.5\linewidth]{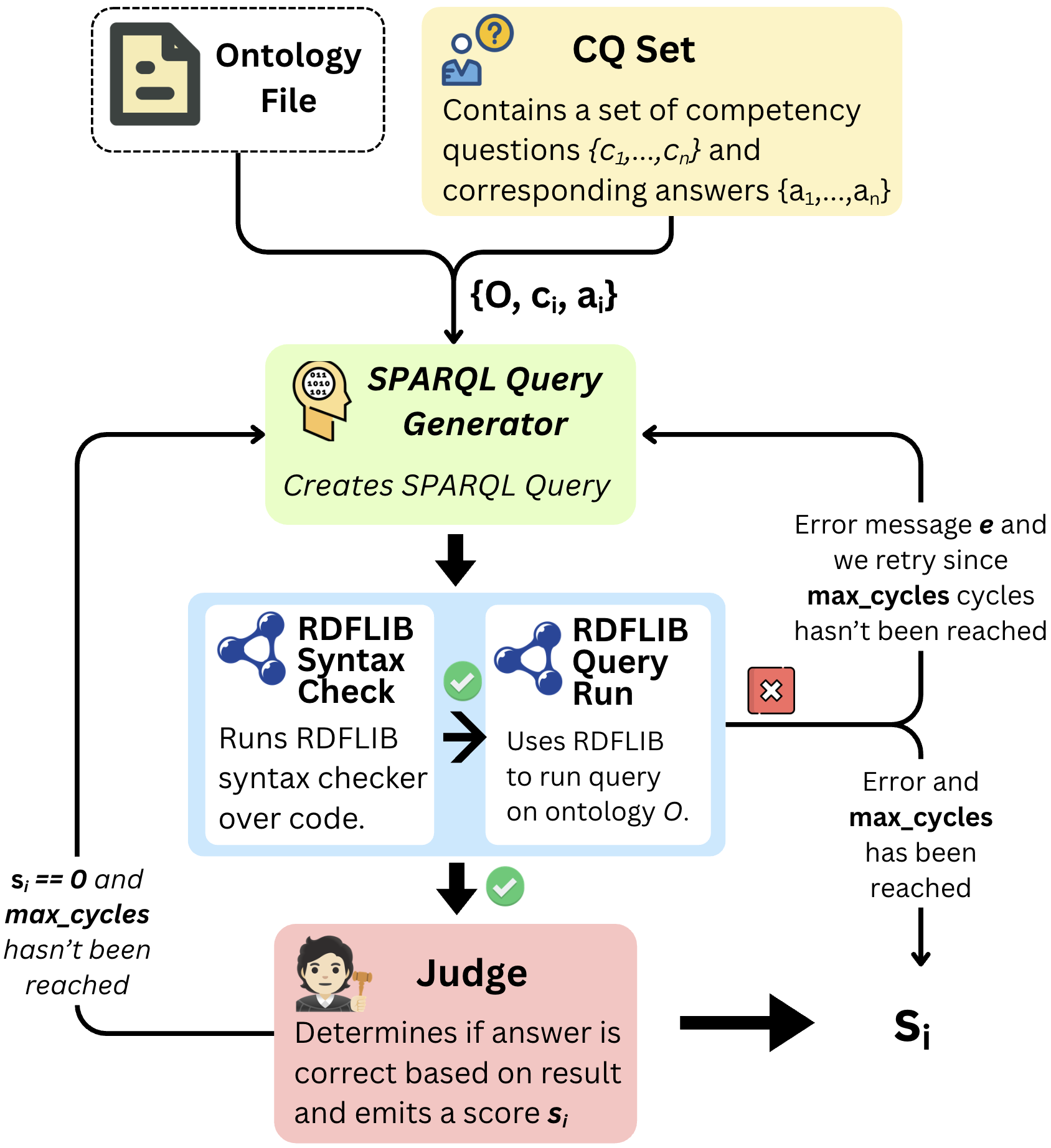}
    \caption{Workflow of the Evaluation Framework}
    \label{fig:evaluation_method}
\end{figure}

The evaluation framework is designed to simulate the iterative query formulation process that human practitioners follow. Instead of relying on a single, static SPARQL query per CQ, this system allows a query agent to adapt its strategy based on diagnostic feedback. This approach goes beyond a simple pass/fail metric to measure the pragmatic usability of an ontology.
Figure \ref{fig:evaluation_method} depicts the workflow of the evaluation method.

Our ontology generation task is fundamentally schema-oriented (T-Box), as it aims to construct a conceptual model of domain entities, relationships, and constraints. However, to operationalize CQ evaluation via SPARQL, we introduce synthetic instance data (A-Box), such as \texttt{ex:payment1}, to enable executable query answering.
These instances function as \emph{unit tests} for the schema: they allow us to verify that the modeled classes and properties (e.g., \texttt{ex:hasMonetaryAmount}) are correctly defined and can support query execution. Importantly, the CQs themselves are designed to validate whether the T-Box encodes the necessary relational structure to support \emph{any} future instance data, rather than evaluating the specific instances introduced for testing.

\noindent The evaluation process is implemented as a LangGraph state machine\footnote{\url{https://docs.langchain.com/oss/python/langgraph/graph-api}} with the following nodes:

\begin{itemize}[leftmargin=*,nosep]
    \item \textbf{Query Generation Agent:} The LLM receives the ontology, the natural language CQ, and the expected answer. It produces a SPARQL query, with structured output constraints to ensure syntactic integrity.
    \item \textbf{Syntax Validation Node:} A candidate SPARQL query is parsed to verify compliance with SPARQL syntax rules. If the query is invalid, a feedback message detailing the error is sent back to the agent for regeneration.
    \item \textbf{Query Execution Node:} A syntactically valid query is executed against the in-memory ontology graph. Execution failures are detected, and the error trace is fed back to the agent.
    \item \textbf{Answer Evaluation Node:} When a query executes successfully, a second LLM, the \textbf{Judge Agent}, compares the result to the expected answer from the CQ using a formal rubric. The rubric assigns a score of \textbf{1.0} for a fully correct answer, a fractional score (e.g., \textbf{0.x}) for a partially correct answer that might miss some details from the complete answer, and \textbf{0.0} for an incorrect or irrelevant result. If the score is 0.0, the judge's justification is used to guide query refinement.
    \item \textbf{Failure Handler Node:} If the maximum number of refinement cycles is reached without a satisfactory answer, the evaluation is finalized with a low score.
\end{itemize}

The iterative nature of this system is central to its realism. It mirrors how a human user might refine a query based on execution errors or unexpected results, with the goal of measuring not only whether the ontology can answer a CQ but also how accessible that answer is. An ontology that requires minimal refinement cycles to produce a correct answer is considered more usable in practice.

For example, for the CQ `What is the monetary amount the beneficiary is entitled to?', the Query Generation Agent initially proposes a simple query such as the one provided in \Cref{lst:initial-query}.

\begin{minipage}{\textwidth}
\begin{lstlisting}[style=codeblock, label=lst:initial-query, caption={Initial Query}]
SELECT ?amount WHERE {
  ex:payment1 ex:hasAmount ?amount .
}
\end{lstlisting}
\end{minipage}
\hfill

This query fails because the property is actually \texttt{ex:hasMonetaryAmount} (and not \texttt{ex:hasAmount}). The framework provides this feedback, and the agent refines the query. After a successful refinement, the final query would look like the one provided in \Cref{lst:final-query}.

Upon execution, the result set `[ { ``value": ``50000", ``currency": ``USD" } ]' is passed to the Judge agent. Given the expected answer `50000 USD', the Judge agent assigns a score of `1.0'.

At the end of an evaluation run, every CQ is annotated with its final query, result, the judge's justification, and a final score. The system computes an average CQ score for the entire ontology, providing a holistic performance metric that is stored in a structured report for further analysis.

\begin{lstlisting}[style=codeblock,label=lst:final-query, caption={Final Refined Query}]
PREFIX ex: <http://www.example.org/ontology#>
PREFIX xsd: <http://www.w3.org/2001/XMLSchema#>

SELECT ?value ?currency WHERE {
  ?payment a ex:PaymentEvent ;
           ex:hasEntitledParty ex:beneficiary1 ;
           ex:hasMonetaryAmount ?monetaryAmount .
  ?monetaryAmount ex:hasValue ?value ;
                  ex:hasCurrency ?currency .
}
\end{lstlisting}

\subsection{Vector-Based Retrieval-Augmented Generation (RAG) Evaluation for Latent Knowledge Assessment}

While the automated SPARQL evaluation framework measures the pragmatic usability of the generated ontologies, it may conflate ontology quality with the querying proficiency of the SPARQL Query Generation Agent. Because SPARQL is highly sensitive to syntax, URI resolution, and vocabulary mismatch, failed queries or empty result sets do not necessarily imply that the relevant knowledge is absent from the ontology; they may instead reflect query formulation failure.

To complement this limitation, we introduce a semantic RAG-based evaluation. Although this method does not test formal OWL semantics directly, it provides a more fault-tolerant measure of knowledge coverage and structural accessibility by evaluating whether relevant ontology nodes and their connections can be successfully retrieved and traversed.

\subsubsection{Semantic Retrieval and Context Expansion}
Our vector-based evaluation draws methodological inspiration from the MINE-2 metric introduced by the KGGen framework~\cite{mo2025kggen} for knowledge graph evaluation. However, rather than evaluating on open-domain datasets like WikiQA~\cite{yang2015wikiqa}, our approach explicitly leverages our domain-specific CQ set.

Our RAG pipeline operates directly on the generated ontologies by first transforming them into a formal directed graph, $G = (V, E)$. Using RDFLib, we extract the complete set of semantic triples $T = \{(s, p, o)\}$. To ensure semantic consistency, we apply a string normalization function $f(x)$ to all triple components. The vertex set $V$ is defined as the union of all normalized subjects and objects:

$$V = \{f(x) \mid \exists (s,p,o) \in T \text{ where } x = s \lor x = o\}$$

Because these components are derived directly from the ontology, $V$ inherently comprises a heterogeneous mix of classes, instances, and property definitions. The edge set $E$ is then constructed using the normalized predicates $f(p)$ to establish directed relationships between these nodes:

$$E = \{(f(s), f(o), f(p)) \mid (s,p,o) \in T\}$$

Finally, we map the graph into a dense vector space by independently encoding the node set $V$ and the edge set $E$ into separate variables using the \texttt{all-MiniLM-L6-v2} embedding model\footnote{\url{https://huggingface.co/sentence-transformers/all-MiniLM-L6-v2}}.

To retrieve the most relevant ontological context for a given CQ, we initially designed a hybrid search strategy. For a query $q$ and an ontological element $e$, the retrieval score was computed as a weighted linear combination of sparse lexical matching and dense semantic similarity ~\cite{mo2025kggen}:

\begin{equation}
Score(q, e) = \alpha \cdot BM25(q, e) + \beta \cdot CosineSimilarity(q, e)
\end{equation}

The weights $\alpha$ and $\beta$ were explored through iterative experimentation. We found that BM25(Best Matching 25) ~\cite{bm25okapi}, a probabilistic ranking function based on term-weighting, introduced significant noise in this domain, rendering term-frequency signals unreliable for ontology element identifiers. Consequently, our final configuration assigns $\alpha = 0$ and $\beta = 10.0$, relying exclusively on dense semantic similarity. This approach, combined with our decoupled node/edge indexing, empirically yielded higher evaluation scores by effectively stripping away lexical noise.

Because CQs often require multi-relational reasoning, retrieving isolated nodes or edges is insufficient. Upon identifying the top-$k$ most relevant elements, the system performs a localized graph traversal, expanding outward from each retrieved element to a depth of two hops. This aggregates all connected subject-predicate-object triples into a cohesive, human-readable context block.

\paragraph{RAG as a Structural (Topological) Metric.}
It is important to clarify the scope of the RAG-based evaluation. This method is \emph{not} intended to validate logical correctness under OWL semantics, such as description logic entailment, subsumption, or consistency. Instead, it evaluates \emph{topological coherence} and \emph{knowledge accessibility} within the generated ontology.

Because retrieval operates directly over the ontology graph and requires multi-hop traversal, successful answer synthesis implicitly depends on several structural properties: (i) correct entity linking between relevant concepts, (ii) meaningful and semantically informative edge definitions, and (iii) sufficient graph connectivity to support multi-step reasoning. In this sense, the evaluation acts as a probe of whether the ontology encodes knowledge in a navigable and connected form.

Since multi-hop synthesis is strictly bounded by the graph's topological quality, a structurally deficient ontology (e.g., one consisting of disconnected components or shallow ``star''-like fragments) will cause the traversal process to terminate prematurely, yielding insufficient context for answer synthesis. This ensures that the RAG score reflects structural modeling integrity and relational coherence, rather than merely lexical or embedding-based similarity.

Therefore, we position the RAG-based evaluation as a complementary metric to SPARQL-based reasoning: while SPARQL assesses formal queryability under strict semantics, RAG provides a more fault-tolerant measure of whether the ontology captures and connects the underlying domain knowledge in a practically usable form

\subsubsection{LLM-as-a-Judge Evaluation}
\label{sec:llm-as-a-judge-evaluation}

With the localized subgraph retrieved, we employ the DSPy framework~\cite{khattab2023dspy} to automate the answering and scoring process. The process is executed in two stages:
\begin{enumerate}
    \item \textbf{Answer Synthesis:} An LLM is prompted with the retrieved context and the original CQ, tasked with synthesizing a one-sentence answer solely based on the provided ontological facts. 
    \item \textbf{Automated Adjudication:} A secondary LLM-as-a-Judge compares the synthesized answer against the ground-truth expected answer derived during the initial CQ generation phase. 
\end{enumerate}

The judge utilizes a CoT \cite{wei2023chainofthoughtpromptingelicitsreasoning} reasoning path to assign a granular score: $1.0$ if the synthesized answer is fully correct and semantically equivalent to the expected answer, $0.5$ if it is partially correct (the answer may have most of the details but miss a detail or two), and $0.0$ if it is incorrect.

By utilizing this vector-based RAG method, we establish a robust upper-bound measurement of the ontology's informational fidelity. It provides a fault-tolerant metric that proves the LLM successfully extracted and modeled the necessary domain knowledge, even in instances where the strict logic of a SPARQL agent failed to retrieve it.

To mitigate potential risk of evaluation circularity arising from the use of LLMs across multiple stages of the pipeline, including ontology generation, query formulation, and answer adjudication, we adopt several safeguards. First, architectural quality evaluation is performed using a panel of heterogeneous, state-of-the-art LLMs from different model families, reducing the likelihood of single-model bias dominating the results. Second, we enforce strict separation between generation and evaluation prompts, ensuring that the judging process operates independently of the generation context. 

\section{Results and Discussion}
\label{sec:results}

We evaluated both generation approaches across two primary dimensions: architectural fidelity and functional coverage. Architectural fidelity was measured using a panel of five state-of-the-art LLMs (Gemini 3.1 Pro, Claude Sonnet 4.6, Grok 4.1, Deepseek-V3.2, Meta Llama 4) acting as expert judges to score the generated ontologies on a 1-to-5 scale across three key metrics: 

\begin{enumerate}
    \item \textbf{Extensibility:} Evaluates whether the enforced use of patterns like \emph{Event Reification} and \emph{Participation} yields a flexible ontology capable of supporting future domain growth.
    
    \item \textbf{Redundancy:} Assesses the strict avoidance of duplicated classes, properties, and rules by measuring the model's success in reusing existing ontology components over generating overlapping entities.
    
    \item \textbf{ODP Usage:} Quantifies adherence to established ODPs to ensure the output aligns with semantic engineering best practices rather than ad-hoc LLM invention.
    
\end{enumerate}

Functional usability was evaluated using our automated SPARQL and RAG framework, which computes the average CQ score achieved by each ontology. We evaluated two publicly available, synthetically generated life insurance contracts---Equivita and Sentinel---summarized in \Cref{tab:contract_comparison}. We intentionally restrict the study to two contracts to enable controlled, deep diagnostic analysis rather than broad benchmarking.

Both contracts are 20-year level term life insurance policies and share a common structural backbone, including core provisions such as benefits, death benefit clauses, and contestability terms. However, they differ meaningfully in modeling complexity. Equivita includes more legal depth, with features such as simultaneous death handling, irrevocable beneficiaries, and conversion-at-expiry rights. Sentinel is more event- and constraint-heavy, primarily due to its \emph{Accelerated Death Benefit} rider, which introduces nested eligibility conditions, an additional event structure, and new financial and temporal constraints, such as a 50\% payout cap and terminal illness certification.

Taken together, the two contracts provide a moderate but realistic ontology-engineering testbed: they include multi-party relationships, overlapping temporal constraints, and financial conditions that must be modeled as structured entities rather than flat predicates. Each contract was evaluated with roughly 150 CQs, consisting mostly of single-hop questions with some multi-hop cases.

\begin{table}[t]
    \centering
    \renewcommand{\arraystretch}{1.2}
    \caption{Comparison of the two insurance contracts used in evaluation.
    }
    \label{tab:contract_comparison}
    \begin{tabular}{|p{5cm}|p{4.8cm}|p{4.8cm}|}
        \hline
        \textbf{Metric} & \textbf{Contract 1 (Equivita)} & \textbf{Contract 2 (Sentinel)} \\ \hline
        \textbf{Insurer} & Equivita Mutual Life Insurance & Sentinel Life Assurance Co. \\ \hline
        \textbf{Policy Type} & 20-Year Level Term & 20-Year Level Term \\ \hline
        \textbf{Word Count} & 2331 & 1953 \\ \hline
        \textbf{Major Clauses} & 9 & 8 \\ \hline
        \textbf{Defined Terms} & 13 & 10 \\ \hline
        \textbf{Named Roles/Parties} & 3 & 5 \\ \hline
        \textbf{Riders Attached} & None & 1 (Accelerated Death Benefit) \\ \hline
        \textbf{Governing Law} & New York & Colorado \\ \hline
        \textbf{Number of CQs Evaluated} & 161 & 149 \\ \hline
    \end{tabular}
\end{table}

\subsection{Ontological Quality and Architectural Fidelity}

\begin{table}[htbp]
\centering
\caption{Average Quality Metrics for Contract 1 (Equivita) and Contract 2 (Sentinel) Life Insurance Contracts}
\label{tab:quality-metrics}
\small 
\begin{tabular}{llcc}
\hline
\textbf{Sample} & \textbf{Metric} & \textbf{Direct Generation} & \textbf{Multi-Agent} \\ \hline
\multirow{4}{*}{Equivita} & Extensibility & 1.60 & \textbf{3.80} \\
 & Redundancy & 1.00 & \textbf{2.40} \\
 & ODP Usage & 1.40 & \textbf{3.80} \\
 & \textit{Overall} & 1.33 & \textbf{3.33} \\ \hline
\multirow{4}{*}{Sentinel} & Extensibility & 2.80 & \textbf{3.80} \\
 & Redundancy & \textbf{2.00} & 1.80 \\
 & ODP Usage & 2.20 & \textbf{3.40} \\
 & \textit{Overall} & 2.33 & \textbf{3.00} \\ \hline
\end{tabular}
\end{table}

\begin{table}[t]
    \caption{Inter-rater Reliability measured using Krippendorff's alpha for each evaluation metric.}
    \centering
    \renewcommand{\arraystretch}{1.2}
    \begin{tabular}{|l|c|}
        \hline
        \textbf{Metric} & \textbf{Krippendorff's Alpha} \\ \hline
        Extensibility & 0.784 \\ \hline
        Redundancy & 0.537 \\ \hline
        ODP Usage & 0.706 \\ \hline
        \textbf{Overall} & \textbf{0.755} \\ \hline
    \end{tabular}
    \label{tab:krippendorff_alpha}
\end{table}

The multi-agent framework demonstrated a clear superiority in producing structurally sound and standard-compliant ontologies. As shown in Table \ref{tab:quality-metrics}, the multi-agent pipeline achieved the highest overall quality score of 3.33 for Equivita and 3.00 for Sentinel, compared to baseline scores of 1.33 and 2.33 respectively for the direct generation approach.
We further observe the following:

\begin{itemize}[leftmargin=*]
    \item \textbf{ODP Usage and Extensibility:} The most significant improvement was observed in ODP Usage, where the multi-agent pipeline for the Equivita contract scored 3.80, more than double the baseline score of 1.40. This validates the Manager agent's TIP as the primary quality driver, a finding we analyze in depth in \Cref{sec:architectural-quality-and-functional-coverage}. Computationally, this decoupling succeeds by externalizing the logical design. By generating reasoning tokens upfront in a plain-text plan of classes and ODPs, the Coder agent dedicates its limited context window primarily for TTL syntax translation. Consequently, the multi-agent approach significantly improved Extensibility (3.80 vs. 1.60), demonstrating that forced adherence to patterns like Event Reification yields a highly flexible, future-proof foundation.
    
    \item \textbf{The Challenge of Redundancy:} Despite improvements in structural design, redundancy remained difficult for all methods. The highest score was only 2.40, achieved by the multi-agent pipeline on Equivita, while the baseline scored 1.00. This suggests that avoiding redundancy is fundamentally harder for LLMs than applying ODPs. Whereas ODP application depends mostly on localized modeling decisions, redundancy requires global comparison across the evolving ontology to identify and reuse semantically overlapping classes and properties. In practice, this broader state-tracking overwhelms the models attention mechanism often causing it to introduce slightly varied concepts instead of reusing existing ones.

    \item \textbf{Inter-rater Reliability:} As shown in \Cref{tab:krippendorff_alpha}, Krippendorff's alpha~\cite{hayes2007answering} was higher for \emph{Extensibility} (0.784) and \emph{ODP Usage} (0.706) than for \emph{Redundancy} (0.537). This lower agreement is consistent with the difficulty of judging redundancy, which requires broader cross-file semantic comparison and leaves more room for subjective interpretation of what counts as unnecessary duplication. The small evaluation pool ($N=4$ artifact units, each scored by 5 distinct raters per metric) further amplifies the effect of isolated disagreements, causing a larger penalty to alpha than would typically occur in a larger sample.
\end{itemize}

\subsection{Architectural Quality and Functional Coverage}
\label{sec:architectural-quality-and-functional-coverage}

\begin{table}[t]
\centering
\caption{CQ Coverage Results for Equivita and Sentinel Contracts}
\label{tab:combined-cq-coverage}
\small
\begin{tabular}{llcc}
\hline
\textbf{Sample} & \textbf{Metric} & \textbf{Direct Generation} & \textbf{Multi-Agent} \\ \hline
Equivita & Processed SPARQL CQ Coverage & \textbf{.6304} & .6196 \\ \hline
Sentinel & Processed SPARQL CQ Coverage & .4027 & \textbf{.4383} \\ \hline
Equivita & RAG CQ Coverage & .5994 & \textbf{.7112} \\ \hline
Sentinel & RAG CQ Coverage & .4450 & \textbf{.5510} \\ \hline
\end{tabular}
\end{table}

As detailed in \Cref{tab:combined-cq-coverage}, the multi-agent framework demonstrated strong performance across both evaluation paradigms. Under the formal SPARQL evaluation, while the baseline method achieved a negligible 1\% edge on the Equivita contract (63.04\% vs. 61.96\%), the multi-agent framework demonstrated a clear advantage on the Sentinel contract (43.83\% vs. 40.27\%). However, these syntactic querying scores require important contextualization to understand the pipeline's true capabilities.

During the automated SPARQL evaluation, we observed a limitation in the LLM-as-a-judge: it occasionally awarded partial credit to syntactically valid queries that returned empty result sets. To ensure that the metric reflected actual knowledge retrieval rather than query-generation proficiency, we post-processed all empty results to a score of $0.0$. As a result, the ``Processed SPARQL'' scores should be interpreted as a conservative floor for functional usability.

Manual inspection further suggests that these scores are deflated by a representation mismatch between the expected answers and the queried outputs. Because the CQs were generated directly from the English contract, the expected answers are often phrased as high-level legal concepts (e.g., \textit{``the Policy''}), whereas SPARQL commonly returns specific ontology instances or URIs (e.g., \texttt{ex:SLAC-2025-001}). In these cases, the judge LLM often failed to recognize that the retrieved instance satisfied the broader expected concept, leading to false penalties despite correct retrieval.

This brittleness is less pronounced under the vector-based RAG evaluation, where the multi-agent framework achieved coverage scores of 0.7112 and 0.5510, compared to 0.5994 and 0.4450 for the baseline. A key reason is that the RAG pipeline synthesizes the retrieved ontology context into a natural-language answer before adjudication. This reduces the representation mismatch that affected SPARQL evaluation and provides a more fault-tolerant estimate of the latent knowledge captured by the ontology.

Ultimately, the consistent superiority of the multi-agent framework across both of these distinct evaluation paradigms highlights a fundamental difference in how the two systems approach ontology generation. This performance gap stems directly from the architectural advantages of structured, artifact-driven planning:

\begin{itemize}[leftmargin=*]
    \item \textbf{The Illusion of Thinking in Generalized Prompts:} In the baseline approach, the generator receives detailed instructions and is explicitly prompted to reason step by step, creating the appearance of deliberate planning. In practice, however, it must still produce the final ontology immediately through structured JSON tool calls, collapsing planning and implementation into a single step. The model must simultaneously interpret the legal text, determine the schema needed to answer the CQs, and produce valid TTL syntax, which often leads to dropped constraints and unanswerable CQs.

    \item \textbf{Externalized Reasoning and Traceability:} The multi-agent system improves on this by separating conceptual planning from syntactic implementation. The Manager's TIP specifies the modeling choices, including a \textit{CQ Alignment} section linking each CQ to the required classes and ODPs. This makes the intended structure explicit and traceable before coding begins. As a result, the Coder can focus on implementing the plan in valid TTL, rather than simultaneously designing the ontology and encoding it.
\end{itemize}

\subsection{Limitations in Iterative Remediation and Instruction Following}
While the multi-agent architecture yields a superior final artifact, observing the pipeline's execution reveals significant challenges in how LLMs handle iterative debugging and context management. 
The data suggests that the framework's success is heavily reliant on front-loaded planning rather than its downstream QA mechanisms. 

\begin{itemize}[leftmargin=*]
    \item \textbf{The Computational Reality of the Bug-Fixing Loop:} The architecture establishes a safety circuit-breaker of 1,000 maximum cycles to prevent infinite loops. However, this limit is not a functional target, as a single debugging cycle is highly computationally expensive. To execute one cycle, the LLM must ingest the error message, generate a tool call to read the file, ingest the returned textual context, formulate a conceptual fix, generate a tool call to write the fix, and finally wait for the ontology to be re-validated by the RDF parser or HermiT reasoner ~\cite{Shearer2008HermiTA}. In practice, we observed three distinct execution trajectories: (1) the ontology is generated flawlessly on the first pass, circumventing the bug fixer entirely (the most common successful scenario); (2) the pipeline encounters a minor syntax error, struggles through approximately 1 to 50 expensive, prolonged cycles, and eventually patches it; or (3) the agent enters an unrecoverable hallucination loop where it continuously fails to resolve the error until the run is manually aborted.
    
    \item \textbf{Pathologies in Autonomous Remediation:} When the pipeline falls into an unrecoverable loop, manual inspection of the agent's tool-use logs reveals several recurring, systemic pathologies:
    \begin{itemize}
        \item \textit{Improper Tool Usage:} Despite explicit system instructions to expand its search window (e.g., reading 200 lines before and after an error), the agent frequently fixates on a narrow range around the exact line number reported by the syntax parser. It repeatedly tries to read lines from the file on the same narrow segment, failing to realize that the root cause (such as a missing period) occurred dozens of lines prior.
        \item \textit{State Amnesia and Failed Meta-Reasoning:} To manage context window limits and prevent out-of-memory errors, the pipeline clears the bug-fixer's raw execution history (its specific tool calls and exact syntactic edits) after a failed cycle. To compensate for this wiped memory, the architecture triggers a meta-reflection step where the LLM summarizes its past attempts in plain text before the next cycle begins. However, we observed that this reflection step was largely ineffective at breaking failure loops.
        \item \textit{Syntax Hallucinations and File Bloat:} When the model cannot deduce the correct logical fix, it frequently resorts to syntactic guessing. We observed instances where the agent continuously appended trailing commas, semicolons, or repetitive triples to the end of a block. Rather than resolving the error, this guessing bloated the ontology file size with garbage output.
    \end{itemize}
    
    \item \textbf{Model Size and Context Degradation:} 

    We conduct our experiments using a 30B-parameter class model, which reflects a realistic deployment setting in resource-constrained and privacy-sensitive environments such as legal and financial domains. In such settings, reliance on smaller or locally deployable models is often preferred due to cost, latency, and data governance considerations.
    
    From an experimental perspective, this choice also serves as a \emph{stress test} for the proposed architecture: it allows us to evaluate whether the multi-agent decomposition can compensate for known limitations of mid-scale models, including context window constraints, state tracking issues, and susceptibility to redundancy. The observed failure modes (e.g., context degradation and ineffective reuse of existing ontology components) are therefore indicative not only of model limitations but also of the architectural challenges inherent to ontology generation. We acknowledge that larger or reasoning-enhanced models may alleviate some of these issues, particularly in planning fidelity and long-range dependency tracking. 
    
\end{itemize}

Ultimately, these execution patterns reveal that while LLMs can generate complex syntax in a single forward pass, they struggle to seamlessly transition between generative and corrective tasks within a bloated, multi-turn context window. Therefore, the primary driver of the multi-agent system's higher quality metrics is the strict, front-loaded guidance provided by the Manager’s TIP, rather than the system's ability to iteratively debug and heal itself.

Due to the high computational cost of the multi-agent pipeline, particularly the bounded iterative QA loops and multi-stage artifact generation, we report results from a single run per configuration (method $\times$ contract). While this limits statistical robustness and prevents direct estimation of variance across runs, our primary objective in this work is to conduct a controlled, diagnostic analysis of architectural design choices and their associated failure modes. Finally, we do not report detailed runtime or cost analysis; however, the multi-agent pipeline introduces additional overhead due to iterative QA loops. Quantifying this tradeoff is an important direction for future work.

\section{Conclusion and Future Work}

This paper presented a controlled empirical study of automated ontology generation from unstructured natural language, comparing a direct LLM-based baseline with a coordinated multi-agent architecture under identical conditions. Both systems were guided and evaluated using automatically derived CQs, which served both as structured functional requirements and as the basis for SPARQL and RAG-based evaluation. Our main finding is that artifact-driven planning consistently improves ontology quality. More broadly, the results suggest that upstream architectural planning is a stronger driver of performance than downstream repair.

At the same time, the results show that structural quality and executable queryability remain only partially aligned. Although the multi-agent system improves architectural fidelity, SPARQL CQ coverage remains moderate (approximately 40\%--63\%), indicating that producing ontologies that are both well-structured and reliably queryable under strict formal constraints remains difficult. In this sense, ontology generation from unstructured text is still a dual problem: constructing semantically coherent representations while also ensuring that they support precise formal querying. In this context, the SPARQL results should be interpreted as evidence of the remaining challenges in bridging structural modeling and executable reasoning, rather than as a limitation of the evaluation framework.

Recent reasoning-focused LLMs may appear to reduce the need for explicit multi-agent decomposition by internalizing planning through latent reasoning. However, our goal is not to replace such capabilities, but to externalize them into inspectable artifacts. Intermediate outputs such as the SRD and TIP provide explicit checkpoints for transparency, traceability, auditing, and intervention, which is particularly important in high-stakes domains such as legal and financial contracts. Rather than requiring practitioners to interpret or trust opaque internal reasoning traces, our approach enables direct inspection, validation, and governance of each stage in the reasoning process. In this sense, multi-agent decomposition complements modern reasoning models by providing a structured and auditable interface to their capabilities.

Several challenges remain open. The most immediate are redundancy and autonomous debugging: LLMs still struggle to maintain global semantic consistency across long documents and to recover reliably from iterative syntax-repair loops. One promising direction is to introduce an explicit external class registry to support ontology-wide reuse without overloading the active context window.

Our evaluation framework also leaves important room for improvement. First, it partially conflates schema design with instance-level validation; a cleaner separation between T-Box expressivity and A-Box population would improve future evaluations. Second, LLM-mediated evaluation introduces unavoidable risks of circularity and bias, even with mitigation steps in place. Incorporating Human-in-the-Loop (HITL) review into the QA stage, comparing against curated gold-standard ontologies, and including expert assessment would provide stronger guarantees of correctness and accountability.

Finally, this study prioritized diagnostic depth over repeated-measures experimentation. Although the observed failure patterns---including redundancy, context degradation, and ineffective downstream repair---were consistent across internal development runs, future work should quantify variance across repeated stochastic runs, and perform controlled agent-level ablations. Beyond this, future research should explore the integration of reasoning-native models with latent CoT capabilities into the multi-agent framework, investigating whether stronger internal reasoning can further enhance or partially replace certain agent roles. Applying and stress-testing these approaches across increasingly complex regulatory domains will be the critical next step toward establishing automated ontology engineering as a reliable practice for domain experts outside the traditional Semantic Web community.

\subsection*{Resources}
The contracts samples and our source code are available in our GitHub repository at\\
\url{https://github.com/brains-group/towards_automated_ontology_generation}.

\section*{Acknowledgments}
We acknowledge the support from NSF IUCRC CRAFT center research grant (CRAFT Grant \#22022) for this research. The opinions expressed in this publication do not necessarily represent the views of NSF IUCRC CRAFT. We are also grateful for the advice and resources from our CRAFT Industry Board members in shaping this work.

\bibliography{references}

\appendix

\end{document}